\documentclass[letterpaper, 10pt, conference]{ieeeconf}
\IEEEoverridecommandlockouts    

\usepackage{multirow}
\usepackage{lipsum}
\usepackage{amsmath}
\usepackage{amssymb}
\usepackage{xcolor}
\usepackage{hyperref}
\usepackage[caption=false,font=footnotesize]{subfig}
\usepackage[ruled,vlined]{algorithm2e}
\usepackage{graphicx}
\graphicspath{{./Figures/}}
\newtheorem{definition}{Definition}
\newtheorem{problem}{Problem}
\title{\LARGE \bf
Online Slip Detection and Friction Coefficient Estimation\\ for Autonomous Racing
}

\author{Christopher Oeltjen$^{1*}$, Carson Sobolewski$^{1*}$, Saleh Faghfoorian$^{1*}$, \\Lorant Domokos$^{1}$, Giancarlo Vidal$^{1}$, Sriram Yerramsetty$^{1}$, and Ivan Ruchkin$^{1}$
\thanks{$^{1}$ Trustworthy Engineered Autonomy (TEA) Lab, Department of Electrical and Computer Engineering, University of Florida,
        { \{{\tt\small ca.oeltjen, csobolewski, faghfoorian.m, ldomokos, g.vidal, iruchkin\}@ufl.edu}}}
\thanks{$^{*}$ Equal contribution} 
}

\begin{document}
	
	\maketitle

	\begin{abstract}
		Accurate knowledge of the tire–road friction coefficient (TRFC) is essential for vehicle safety, stability, and performance, especially in autonomous racing, where vehicles often operate at the friction limit. However, TRFC cannot be directly measured with standard sensors, and existing estimation methods either depend on vehicle/tire models with uncertain parameters or require large training datasets. In this paper, we present a lightweight approach for online slip detection and TRFC estimation. Our approach relies solely on IMU and LiDAR measurements and the control actions, without special dynamical or tire models, parameter identification, or training data. Slip events are detected in real time by comparing commanded and measured motions, and the TRFC is then estimated directly from observed accelerations under no-slip conditions. Experiments with a 1:10-scale autonomous racing car across different friction levels demonstrate that the proposed approach achieves accurate and consistent slip detections and friction coefficients, with results closely matching ground-truth measurements. These findings highlight the potential of our simple, deployable, and computationally efficient approach for real-time slip monitoring and friction coefficient estimation in autonomous driving.

	\end{abstract}
	
	\section{Introduction}
	\label{sec:introduction}

    Effective road driving depends on accurate knowledge of key vehicle and environment parameters, among which the tire-road friction coefficient (TRFC) is one of the most critical \cite{LiSong2009,LiYang2015}. 
    The TRFC defines the upper limit of forces that can be transmitted between tires and road, thereby constraining braking, steering, and overall vehicle stability \cite{Nagy2023EGP}. 
    Accurate and real-time estimation of this parameter is essential not only for human drivers, but increasingly for automated decision-making, including trajectory planning in autonomous vehicles \cite{s18030896}. In autonomous racing, operating at or near these friction limits is essential to achieve competitive lap times. Therefore, \textit{online slip detection} is a crucial capability: it provides feedback for exploiting available traction while maintaining stability, and it supports higher-level functions such as adaptive control, trajectory planning, and race strategy. Moreover, since tire conditions strongly influence traction availability, slip detection also plays a key role in monitoring and managing the effective use of tires during high-performance driving \cite{Kalaria2023}.
    
    Unlike velocity or acceleration, however, the TRFC cannot be measured directly by standard onboard sensors. Specialized sensors capable of direct measurement exist; however, their high cost and integration challenges prevent their widespread use. This limitation has motivated extensive research into indirect TRFC estimation methods \cite{uchanski2001road}. 
    
    Broadly, TRFC estimation approaches can be categorized into \textit{cause-based} and \textit{effect-based} approaches. Cause-based approaches aim to characterize road surface properties directly, using cameras, laser scanners, or acoustic sensing \cite{Breuer1992MeasurementOT,Eichhorn1992PREDICTIONAM}. 
    Although these approaches can provide useful surface classification, they typically yield only coarse friction estimates and often require large training datasets. \textit{Effect-based} approaches instead infer TRFC from the dynamic responses of the vehicle, exploiting measurable quantities such as wheel slip, sideslip angle, yaw rate, or acceleration \cite{inproceedings,Singh2015}. These approaches tend to be more accurate, as they incorporate the actual interaction between tire and road, but they tend to rely heavily on vehicle or tire models whose parameters may vary with conditions.
    
    Within effect-based approaches, a large body of research has adopted \textit{model-based} techniques such as Kalman filtering. Extended Kalman Filters (EKF) \cite{inproceedings}, Unscented Kalman Filters (UKF) and Model Predictive Control (MPC) \cite{10752593}, and Cubature Kalman Filters (CKF) \cite{9338557} have all been applied to TRFC estimation, fusing standard vehicle sensors with simplified models. These approaches can significantly improve estimation accuracy, but they remain model-dependent and are sensitive to parameter uncertainty and nonlinearities.
    
    More recently, \textit{model-free} approaches have been explored as alternatives. These approaches aim to bypass parametric modeling by directly mapping sensor signals to friction estimates, and have also been proposed more broadly for vehicle dynamics modeling, such as physics-constrained neural networks for autonomous racing \cite{chrosniak2024deepdynamics}, highlighting the growing interest in model-free and data-driven approaches. Examples include learning-based estimators using neural networks trained on vehicle accelerations, data-driven terrain classification methods, and vision–language model-based clustering for friction estimation in off-road contexts \cite{gao2019multisensor}.
    Such approaches show promise in handling complex, nonlinear interactions without explicit models, but they generally require extensive datasets and may lack generalization to unseen conditions.  
    
    In this paper, we propose a \textit{lightweight} approach for slip detection and TRFC estimation that does not require tire or dynamical models, parameter identification, or large training datasets. Instead, it leverages only standard onboard motion sensors (IMU and LiDAR), making it simple and practical to deploy. We first design a slip detection module, which determines the precise moment when the tires lose grip. At that moment, our friction estimation module computes the maximum friction coefficient directly from the observed vehicle motion.
    
    Despite its lightweight nature, our approach detects the slip of a 1:10-scale autonomous racing car on a closed-loop track with high accuracy and minimal delay. Furthermore, our friction estimation method accurately estimates the tire-road friction coefficient, effectively differentiating between two test surfaces of varying friction.

    This paper makes three contributions: (i)  A lightweight slip detection technique using only IMU and LiDAR readings and control actions; (ii) A real-time estimation method for TRFC without explicit tire/vehicle modeling or training data; (iii) An experimental validation on 1/10-scale racing cars, showing reasonable online performance of our approach.
    
    The rest of this paper is organized as follows. Section~\ref{sec:problem} provides the key definitions and formulates the slip detection and friction estimation problems. Section~\ref{sec:related_works} reviews the related work. Section~\ref{sec:approach} presents the proposed lightweight approach, and Section~\ref{sec:results} reports its experimental validation. Finally, Section~\ref{sec:conclusion} concludes the paper.

	\section{PROBLEM FORMULATION}
    \label{sec:problem}
    This section formalizes the key concepts and notation underlying our approach. We interleave general definitions and the specifics of autonomous racing, following standard notions in vehicle dynamics
    by Gillespie \cite{gillespie1992fundamentals} and
    Rajamani~\cite{Rajamani2012}.
    
    \subsection{Vehicle Level}
    
\begin{figure}
        \centering
        \includegraphics[width=0.7\linewidth]{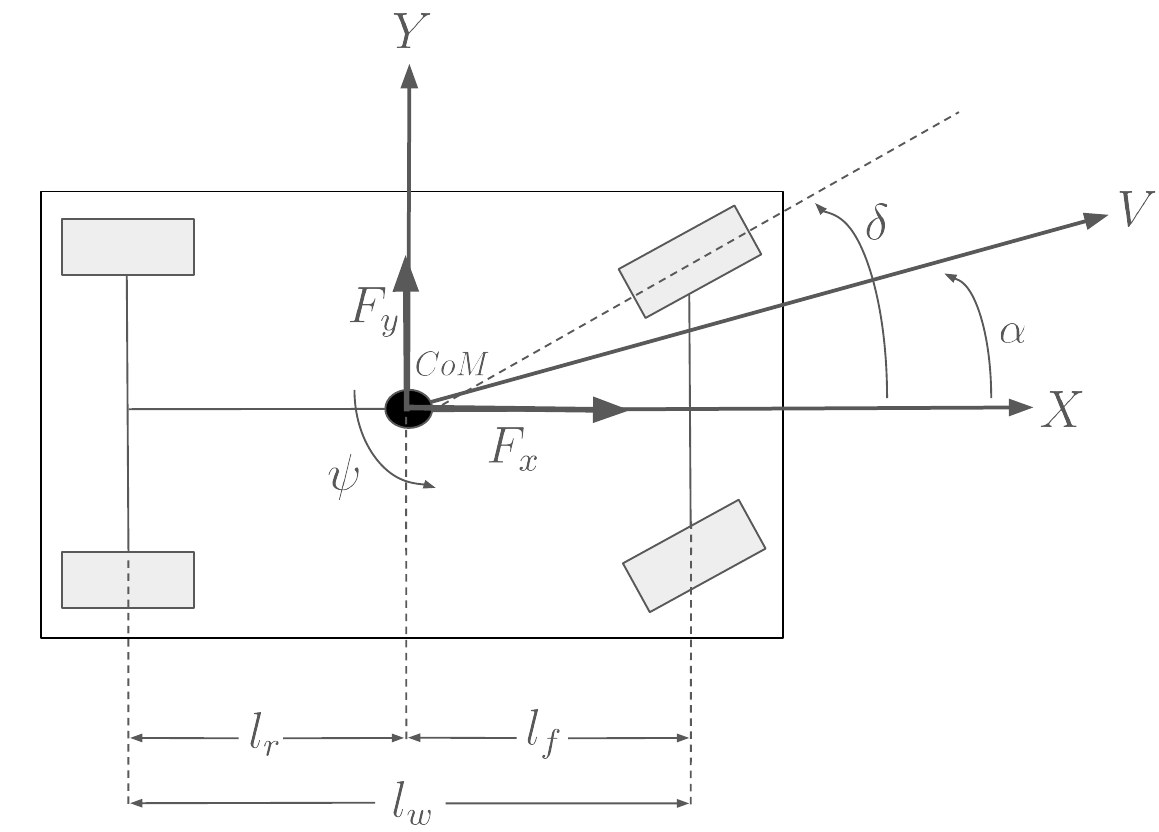}
        \caption{Top view of the vehicle plant $P$ with a body-fixed 3D coordinate system $\{X,Y,Z\}$ attached at the center of mass. The figure illustrates the $X$-$Y$ plane, where $l_f$ and $l_r$ denote the distances from the CoM to the front and rear axles, and $l_w = l_f + l_r$ is the wheelbase. Forces $F_x$ and $F_y$ act at the CoM, $\delta$ is the steering angle, $\alpha$ is the slip angle, and $\psi$ is the yaw angle.}
    
        \label{fig:vehicle}
    \end{figure}

    \begin{definition}[Plant]
        The \textit{plant} $P$ is the physical system under consideration, including its geometry and physical parameters.
        The plant has a center of mass (CoM), where the total mass $m$ is assumed to be concentrated.
        A right-handed body-fixed 3D coordinate system $B=\{X,Y,Z\}$ is attached at point $CoM$, as shown in Figure~\ref{fig:vehicle} (Longitudinal axis $X$ points forward along the chassis, lateral axis $Y$ points leftward at 90\textdegree ~to axis $X$, and the normal axis $Z$ points upward).
    \end{definition}
    
    In our case, the plant is a 1/10-scale autonomous racing car (also referred to as the vehicle), and it is characterized by a set of \textit{configuration}, \textit{inertial}, and \textit{geometric} properties. 
    The configuration is defined by four tires in contact with the road.
    Each tire has an effective tire radius $r_e$, defined as the distance from the tire center to the contact point as shown in Figure~\ref{fig:pure_rolling}.
    The vehicle is rear-wheel drive, while the front axle is actuated (steerable).  
    
    The inertial properties of the vehicle include its total mass $m$ and the yaw moment of inertia $I_z$ about the normal axis $Z$.  
    The geometric properties of our vehicle, illustrated in Figure~\ref{fig:vehicle}, consist of the distances from the CoM to the front and rear axles, denoted by $l_f$ and $l_r$, respectively.

    \begin{definition}[State]
    \looseness=-1
       The time-variant status of plant $P$ is characterized by $n$ state variables $\mathcal{V}$. The state space $\mathcal{S} \subseteq \mathbb{R}^n$ is the set of all feasible states (value vectors) $s$ of variables $\mathcal{V}$.
    \end{definition}
    
    In our case, the state space $\mathcal{S} \subseteq \mathbb{R}^6$, 
    consists of the longitudinal and lateral positions $(x,y)$ expressed in the coordinate system $B$, the yaw angle $\psi$ about the $Z$-axis, the longitudinal and lateral velocities ($v_x$, $v_y$) in $B$, and the yaw rate $\omega_\psi$. Thus, the state vector is
    \begin{equation}
        s = \begin{bmatrix}
            x & y & \psi & v_x & v_y & \omega_\psi
        \end{bmatrix}^\top
    \end{equation}

    \begin{definition}[Sensors]
        For plant $P$ with state space $\mathcal{S} \subseteq \mathbb{R}^n$ and observation space $\mathcal{Y} \subseteq \mathbb{R}^p$, the sensors are represented as function
            $h: \mathcal{S} \to \mathcal{Y} $.
    \end{definition}

    \looseness=-1
    In our case, the observation space $\mathcal{Y} \subseteq \mathbb{R}^5$ includes the outputs of two sensors: (i) an Inertial Measurement Unit (IMU) measures CoM's longitudinal acceleration $\hat{a}_x$ and lateral acceleration $\hat{a}_y$; (ii) a LiDAR measures the CoM's longitudinal velocity $\hat{v}_x$, lateral velocity $\hat{v}_y$, and the yaw rate $\hat{\omega}_\psi$. The IMU measurements were calibrated to compensate for bias, scale factor errors, and axis misalignment, yielding corrected acceleration estimates used in the state observation. The LiDAR-based velocity estimates were obtained using the standard sensor configuration, without additional calibration adjustments.

    Formally, the observation vector $y \in \mathcal{Y}$ is
    \begin{equation}
        y=\begin{bmatrix}
        \hat{a}_x & \hat{a}_y & \hat{v}_x & \hat{v}_y & \hat{\omega}_\psi
    \end{bmatrix}^\top
    \end{equation}
    
    \begin{definition}[Controller]
        For plant $P$ with observation space $\mathcal{Y} \subseteq \mathbb{R}^p$ and action space $\mathcal{U} \subseteq \mathbb{R}^m$, a controller $C$ is a function 
        $q: \mathcal{Y} \to \mathcal{U}$.
        \label{def:controller}
    \end{definition}
    
    Our action space $\mathcal{U} \subseteq \mathbb{R}^2$ contains a control vector $u \in \mathcal{U}$ equal to $     u=\begin{bmatrix}
        v & \delta
        \end{bmatrix}^\top$, coming from a human or software: desired longitudinal velocity $v$ and desired steering angle $\delta$.

    \begin{definition}[Kinematic model]
        For plant $P$ with state space  $\mathcal{S} \subseteq \mathbb{R}^n$ and action space $\mathcal{U}$, a continuous-time kinematic model is a transition map 
        $f: \mathcal{S} \times \mathcal{U} \to \mathbb{R}^n$. 
    \end{definition}
    
    In this paper, we adopt the single-track bicycle model, where the two front wheels are
    lumped into one front wheel, and the two rear wheels into one rear wheel. The corresponding state-transition dynamics are given by
    \begin{equation}
        \dot{x} = v \cos(\psi + \beta), \quad
        \dot{y} = v \sin(\psi + \beta), \quad
        \dot{\psi} = \frac{v}{l_r} \sin(\beta)
        \label{eq:bicycle}
    \end{equation}

    where $\beta$ is the geometric slip angle:
    \begin{equation}
        \beta = \arctan\!\left(\frac{l_r}{l_f+l_r} \tan(\delta)\right)
        \label{eq:_bicycle_slip_angle}
    \end{equation}

    \begin{definition}[Force]
        A \textit{force} $F \in \mathbb{R}^3$ is a vector in coordinate system $B$, representing an interaction between bodies, and consists of three components $(F_x, F_y, F_z)$, where the longitudinal force $F_x$ acts along the forward $X$-axis of the body (positive when pushing forward), the lateral force $F_y$ acts along the $Y$-axis (positive when pushing to the left), and the normal force $F_z$ acts along the $Z$-axis (positive when pushing upward).
    \end{definition}

    \begin{definition}[Dynamical model]
        For plant $P$ with coordinate system $B$, state space $\mathcal{S} \subseteq \mathbb{R}^n$, action space $\mathcal{U} \subseteq \mathbb{R}^m$, and the set of external forces $F$ expressed in $B$, a continuous-time dynamical model is a transition map 
        $g: \mathcal{S} \times \mathcal{U} \times \mathcal{F} \to \mathbb{R}^n$.
    \end{definition}

        Under the bicycle model, subscripts f and r represent the \emph{front} and \emph{rear} axles, respectively, the set of forces $F  = \{F_{x_f}, F_{y_f}, F_{x_r}, F_{y_r}, F_{drag}\}$ consists of tire forces, rolling resistance force $F_{rr}$, and aerodynamical force $F_{drag}$. The vehicle dynamics in the $X_tY_t$ plane in $B$ are given by:
           \begin{align*}
            m\dot{v}_x &= F_{x_f} \cos\delta - F_{y_f} \sin\delta + F_{x_r} - F_{rr} - F_{drag} + v_y \dot{\psi}\\
            m\dot{v}_y &= F_{y_f} \cos\delta + F_{x_f} \sin\delta + F_{y_r} - v_x \dot{\psi} \\
             I_z \dot{\omega}_\psi &= l_f(F_{y_f} \cos\delta + F_{x_f} \sin\delta) - l_r F_{y_r}
       \end{align*}
        
    \looseness=-1
    This paper, however, does not employ any dynamical model.

    \subsection{Tire Level}

\begin{figure}
        \centering
        \includegraphics[scale=0.3]{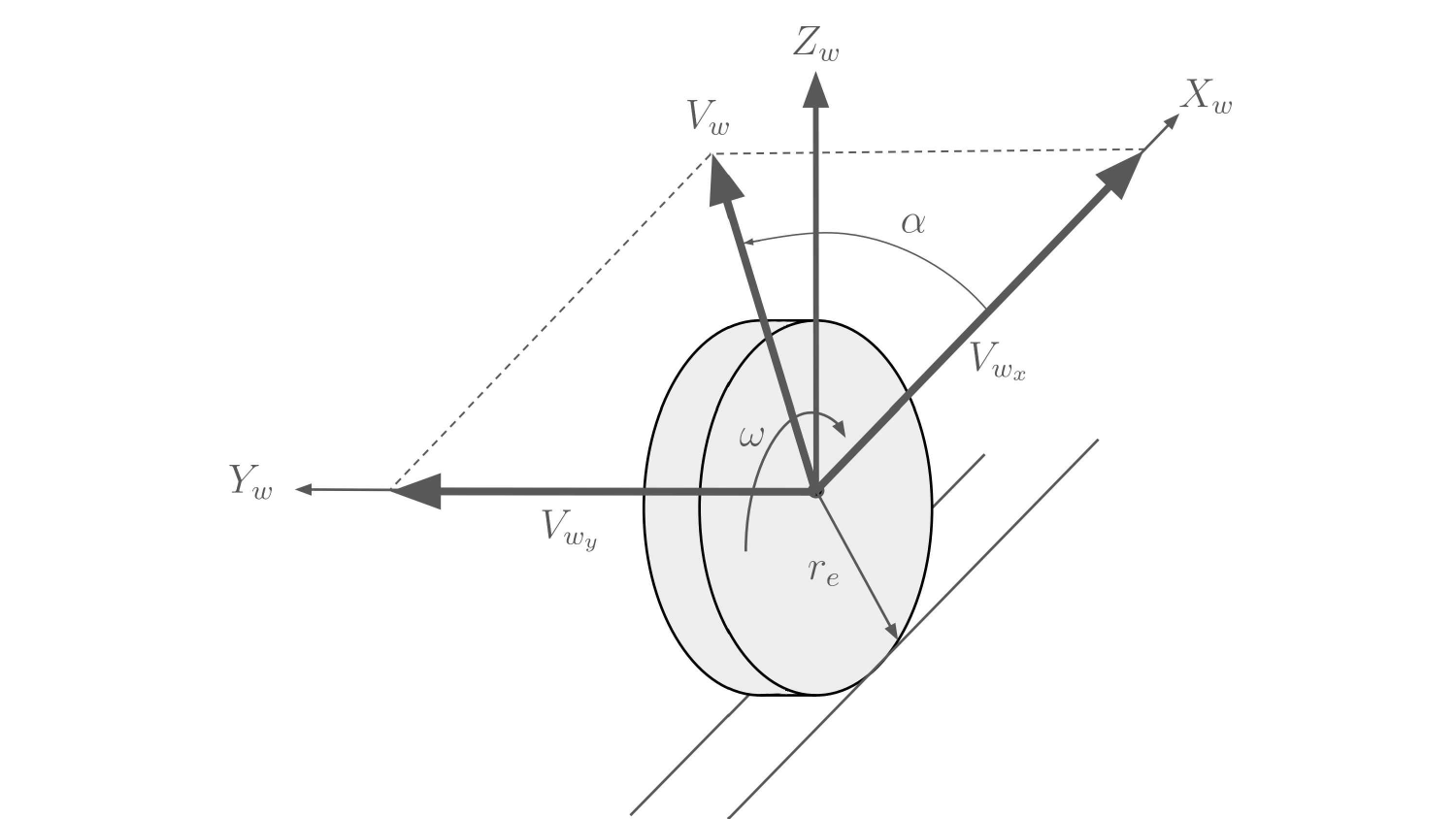}
       \caption{Local body-fixed 3D coordinate system $B_w=\{X_w,Y_w,Z_w\}$ attached at the   tire center.
        The slip angle $\alpha$ is defined as the angle between the tire longitudinal axis $X_w$ and the tire center velocity vector $V_w$.}
    \vspace{-3mm}
        \label{fig:pure_rolling}
    \end{figure}

    \looseness=-1
    Given plant $P$ with coordinate system $B = \{X, Y, Z\}$, we attach a local body-fixed 3D coordinate system $B_w = \{X_w, Y_w, Z_w\}$ to the center of each tire. The $X_w$ axis points in the tire's rotating direction, the $Y_w$ axis points leftward at 90\textdegree ~to the $X_w$ axis, and the $Z_w$ axis is parallel to $Z$ (see Figure~\ref{fig:pure_rolling}). 
    
    The tire has the velocity vector $V_w=(V_{w_x}, V_{w_y})$ in the $X_wY_w$ plane, the angular velocity $\omega$ about the $Y_w$ axis, and the tire-road force is defined as $F_w = \begin{bmatrix}
        F_{w_x} & F_{w_y} & F_{w_z}
    \end{bmatrix}^\top$ in $B_w$.
    The tire has the state space $\mathcal{S}_w \subseteq \mathbb{R}^3$ with the state vector $s_w \in \mathcal{S}_w$ described by $s_w = \begin{bmatrix}
        V_{w_x} & V_{w_y} & \omega
    \end{bmatrix}^\top$.

    \begin{definition}[Pure rolling] 
    A tire with state $s_w=\begin{bmatrix}
        V_{w_x} & V_{w_y} & \omega
    \end{bmatrix}^\top$ and effective radius $r_e$ satisfies the pure rolling condition if:
        \[
        V_{w_x} = r_e \omega, \quad
        V_{w_y} = 0
        \]
    \end{definition}
    \vspace{0.2cm}
    
    \begin{definition}[Slip]
    A tire with state $s_w=\begin{bmatrix}
        V_{w_x} & V_{w_y} & \omega
    \end{bmatrix}^\top$ is in the \textit{slip condition} when the state does not satisfy the pure-rolling condition. Equivalently, slip occurs when at least one of these two scalars is non-zero:
        \begin{itemize}
            \item \textit{Slip ratio:} The slip ratio $\kappa$ is the relative difference between the tire's circumferential speed ($r_e\omega$) and the longitudinal speed of the tire center ($V_{w_x}$)
            \begin{equation}
                \kappa = \frac{r_e\omega - V_{w_x}}{\max(r_e\omega, V_{w_x})}
            \end{equation}
    
            \item \textit{Slip angle:} The slip angle $\alpha$ is between the tire longitudinal axis $X_w$ and the tire center velocity vector $V_w$ as shown in Figure~\ref{fig:pure_rolling}.
            \begin{equation}
                \alpha = \arctan(\frac{V_{w_y}}{V_{w_x}})
            \end{equation}
            
        \end{itemize}
    \end{definition}
    
    \vspace{0.2cm}

    \begin{definition}[Traction coefficient]
       
        For a tire with force $F_w = (F_{w_x}, F_{w_y}, F_{w_z})$ in $B_w$, the traction coefficient $\rho(F_w)$, alternatively called normalized traction force, is defined as
        \begin{equation}
            \rho(F_w) =\frac{\sqrt{F_{w_x}^2+F_{w_y}^2}}{F_{w_z}}
        \end{equation}
    \end{definition}
    \vspace{0.2cm}

    \begin{definition}[Tire Road Friction Coefficient]
        For a given tire $w$, friction coefficient $\mu$ is the maximum value of the traction coefficient $\rho$ for all possible forces $F_w$ under the pure rolling condition:
        \begin{equation}
            \mu = \max_{F_w} \{ \rho(F_w) \;|\; \kappa = 0, \; \alpha = 0 \}
        \end{equation}
    \end{definition}
    \vspace{0.2cm}
    
    For a given normal force $F_{w_z}$ on the tire, the tire-road friction coefficient $\mu$ depends on the characteristics of the road surface, and it constrains the maximum force that a tire can supply in the $X_wY_w$ plane \cite{5510758}.

    \subsection{Problem Statement}
    
    \begin{problem}[Slip detection]
        For unknown vehicle state signal $s(t)$ with the earliest slip at $t^\ast$, given control signal $u(t)$ 
        observation signal $y(t)$,
        and a kinematic model with transition map $f$,
        detect the slip at the earliest time $\hat{t} \ge t^\ast$.    
       
    \end{problem}

    \begin{problem}[Friction coefficient estimation]
        In the context of an unknown state signal $s(t)$ with the true friction coefficient $\mu^\ast$, given an action signal $u(t)$ and an observation signal $y(t)$, 
        estimate the friction coefficient $\hat{\mu}$.

    \end{problem}

    In our setup, the action signal $u(t)$ corresponds to the commanded longitudinal velocity $v$ and steering angle $\delta$, while the observation signal $y(t)$ consists of IMU-based accelerations $\hat{a}_x$ and $\hat{a}_y$, and LiDAR-based odometry (longitudinal and lateral velocities $\hat{v}_x$ and $\hat{v}_y$, and yaw rate $\hat{w}_\psi$). These signals form the basis for both slip detection and friction coefficient estimation under the following \textit{assumptions:}
    \begin{itemize}
        \item The vehicle is a rigid body.
        \item IMU and LiDAR are mounted at the vehicle's CoM.
        \item The road is flat --- no pitch or roll dynamics.
        \item Vertical motion is negligible for both vehicle and tire levels --- planar dynamics only.
        \item No parametric models are learned.
        \item A single-track kinematic model with fixed geometric parameters (wheelbase, axle distances) is assumed.
        \item Aerodynamic drag ($F_{drag}$) and rolling resistance ($F_{rr}$) are negligible. 
        \item The forces are considered at the vehicle level (averaging over the tires), not each individual tire.
    \end{itemize}

	\section{Related Works}
	\label{sec:related_works}
    In classical mechanics, the friction coefficient is defined as the dimensionless ratio 
    between tangential and normal forces at the interface of any two contacting surfaces  \cite{popov2010contact, beer2016vector, bowden2001friction}. In this context, the friction coefficient depends on material properties, surface roughness, and contact conditions. 
    This paper considers the tire–road friction coefficient (TRFC), representing the limit on the combined longitudinal and lateral tire forces in vehicle dynamics. TRFC is often represented by the friction circle, which describes the limit of possible tire force combinations in the horizontal plane.

    \looseness=-1
    Many approaches have been proposed for estimating the friction coefficient~\cite{khaleghian2017technical}. They can be broadly split into two categories: \textit{cause-based} and \textit{effect-based}.

    Cause-based approaches measure friction-related parameters of the road/tire and correlate them with friction in several ways. \textit{Tread sensing} monitors deformations at the tire–road interface to infer friction \cite{Eichhorn1992PREDICTIONAM} \cite{Breuer1992MeasurementOT}. \textit{Optical methods} use cameras or laser-based sensors to detect surface properties such as texture, water, or ice, sometimes employing infrared or polarization measurements, and neural networks trained on image features \cite{andersson2007road} \cite{tuononen2008optical1}. 
    \textit{Acoustic methods} instead classify the road type from tire noise or microphone signals, often combined with machine learning techniques such as support vector machines or neural networks 
    \cite{alonso2014wet} \cite{kongrattanaprasert2009automatic}. 
    While these approaches bypass the need for vehicle or tire models, they require heavyweight hardware instrumentation and are sensitive to environmental disturbances, which limits their deployment in practice. In contrast, our approach relies only on the standard onboard vehicle sensors.

    Effect-based approaches can be split into two main categories: \textit{model-based} and \textit{model-free} approaches. The former aim to estimate TRFC using simplified mathematical or dynamical representations of the tire–vehicle system, typically without requiring additional external sensors. These approaches are generally more accurate and repeatable than cause-based approaches, and can themselves be broadly divided into three categories~\cite{khaleghian2017technical}. First, \textit{vehicle dynamics methods} employ simplified dynamical models such as the single-wheel, quarter-car, bicycle, or four-wheel vehicle models, combined with estimation algorithms including recursive least squares, Kalman filters, or sliding mode observers, to infer unmeasured variables such as tire forces, slip ratios, and ultimately the friction coefficient \cite{woo2024elsd,wang2020ackf_trfc,li2024adaptiveSTKF,rajamani2012algorithms,hsiao2011robust}. Second, \textit{tire-based methods} exploit mathematical tire models, such as the Magic Formula, Dugoff, Brush, or LuGre models, which relate tire forces and moments to slip ratio and slip angle. They estimate the maximum tire–road friction by fitting or observing deviations between measured and modeled behavior~\cite{pacejka1992magic,Svendenius2009BrushValidation,fiala1954tiremodel}.
    Finally, \textit{slip-slope based approaches} rely on the relationship between the initial linear region of the $\mu$–slip curve and its saturation level, using the slope at low slip as a predictor of the maximum available friction~\cite{5510758, Wang2004Friction}. Collectively, these methods represent the dominant body of work on TRFC estimation in the literature. 

    \looseness=-1
    Yet, the effectiveness of model-based approaches is fundamentally tied to accurate knowledge of model parameters and tire-road forces, which are often difficult or impractical to obtain. Tire models, for example, require prior calibration of parameters such as cornering stiffness \cite{inproceedings} or Magic Formula coefficients \cite{christ2021timeoptimal,pacejka1992magic}, while vehicle dynamical models depend on quantities like mass, moment of inertia, and external forces \cite{christ2021timeoptimal}. Estimating these values typically involves additional filtering or indirect measurements \cite{10752593}, but even then, parameters such as the moment of inertia can vary significantly with changes in mass distribution, leading to degraded performance. This strong dependency on parameter identification limits the adaptability of model-based methods, motivating the need for model-free alternatives that bypass explicit parameterization altogether.

    \looseness=-1
    More recently, several model-free approaches have been proposed for TRFC estimation. Unlike model-based approaches, these approaches avoid relying on explicit vehicle or tire models and instead use combinations of signals and data-driven techniques. For example, Sadeghi et al. \cite{sadeghi2022maximum} used wavelet transform and neural networks on vertical acceleration signals to predict the maximum tire–road friction coefficient. Midgley et al. \cite{midgley2025model} developed a machine learning framework for heavy vehicles with pneumatic 
    braking systems, showing that neural networks can outperform classical sliding-mode observers. These studies demonstrate the potential of model-free and learning-based approaches to overcome the limitations of parametric vehicle and tire models. 
    
    Unfortunately, most model-free approaches depend on extensive datasets or specialized sensing modalities, which limit their adaptability and generalization \cite{song2018estimating}. In contrast, our work proposes a lightweight formulation that detects slip and estimates friction directly from motion signals, without requiring significant model fitting or large training data.

	\section{DETECTION AND ESTIMATION APPROACH}\label{sec:approach}

    To detect vehicle slip, our framework leverages a lightweight physics-informed comparison between expected vehicle motion --- derived from control inputs --- and measured vehicle motion, estimated from onboard sensor data. Once the slip is detected, we compute the maximum observed traction coefficient. 
    
    \subsection{Slip Detection}
    
    Our approach separates the longitudinal and angular dimensions of slip, as described below. 
    
    \subsubsection{Linear Detection}
    
    Slip along the $X$ axis is identified by comparing the expected velocity $v$, predicted from the commanded throttle by the vehicle’s onboard odometry system, to the measured longitudinal velocity $\hat{v}_x$, obtained from an EKF that fuses IMU measurements with pose estimates from Monte Carlo localization (MCL) \cite{fox1999mcl}. MCL localizes the vehicle within the map using LiDAR, while the EKF incorporates these pose estimates together with IMU data and a parameter-free, omnidirectional kinematic model to provide the full state, including vehicle velocities. This approach provides a velocity estimate that can be directly compared to the commanded input without relying on explicit dynamical or tire-force modeling.

    A boolean indicator $D_{lin}$ is then defined to detect slip based on a threshold hyperparameter $\delta_{lin}$:
    \begin{equation}
    D_{lin}=|v-\hat{v}_x| \geq \delta_{lin}
    \end{equation} 
    
    The hyperparameter $\delta_{lin}$ is calculated offline, with a heuristic based on previously observed sensor readings without labels. Section~\ref{sec:dataproc} describes our cross-validation for the threshold.

    \subsubsection{Angular Detection}
    In the angular domain, slip is detected by comparing the expected angular velocity $\omega_\psi$ of the vehicle with the measured angular velocity $\hat{\omega}_\psi$ estimated from onboard sensor data, similar to longitudinal detection. 
    
    The expected angular velocity $\omega_\psi$ is derived from the kinematic bicycle model $f$ as follows.
    From Eq.~\ref{eq:bicycle}, the yaw rate is given by

        $\dot{\psi} = \frac{v}{l_r} \sin(\beta)$.

     Substituting the expression for $\beta$ from Eq. \ref{eq:_bicycle_slip_angle} 

    into the yaw rate equation, we obtain
    \begin{equation}
        \dot{\psi} = \frac{v}{l_r} \sin\!\left( \arctan\!\left(\frac{l_r}{l_f + l_r} \tan(\delta)\right) \right)=\frac{v}{l_f + l_r}\tan(\delta).
    \end{equation}

    Thus, the expected angular velocity based on control inputs is
    \begin{equation}
    \omega_\psi=\frac{v}{l_w}\tan(\delta) 
    \end{equation}
    
    Boolean indicator $D_{ang}$ determines whether the car has slipped in an angular fashion, based on the threshold $\delta_{ang}$:
    \begin{equation}
        D_{ang}=|\omega_\psi - \hat{\omega}_\psi| \geq \delta_{ang}
    \end{equation}
    
    The hyperparameter $\delta_{ang}$ is calculated offline based on previously observed sensor readings without labels. Section~\ref{sec:dataproc} describes our cross-validation for the threshold.
    
    \medskip
    
    In summary, slip detection relies on monitoring discrepancies between expected and measured vehicle motion, both linear and angular. A slip event is registered via the respective boolean indicators $D_{ang}$ and $D_{ang}$ once either discrepancy exceeds the corresponding threshold.

    \subsection{Friction Estimation}
    By definition, the tire–road friction coefficient $\mu$ is the maximum traction coefficient without slip, as shown in Eq.~\eqref{eq:mu_def}.
    \begin{equation} \label{eq:mu_def}
    \mu = \max_{F_w} \{ \rho(F_w) ;|; \kappa = 0, ; \alpha = 0 \}
    \end{equation}

    \looseness=-1
    At the tire level, $\mu$ is defined in terms of the longitudinal and lateral forces acting on a single tire. However, to avoid building a tire model, we combine the forces at the vehicle level. Then, all tire-road interactions are aggregated into two external forces (longitudinal and lateral) acting at the vehicle's center of mass: one longitudinal ($F_x$) and one lateral ($F_y$). The resulting friction coefficient should therefore be interpreted as the average value across all four tires. As such, we assume a negligible discrepancy between the tires.
    
    At the vehicle level, we consider the external forces acting on the CoM. Applying Newton’s second law, the longitudinal and lateral forces are related to the IMU-measured accelerations as:
    \begin{equation}
        F_x = m \hat{a}_x, \quad F_y = m \hat{a}_y, \quad F_z=mg
    \end{equation}

    Substituting these relations into the traction coefficient yields:
    \begin{equation}
    \rho(F_w) = \frac{\sqrt{\hat{a}_x^2 + \hat{a}_y^2}}{g}
    \end{equation}
    
    In this paper, this quantity is computed continuously during all non-slip timesteps. 
    Slip conditions are encoded by the boolean indicators $D_{lin}(t)$ and $D_{ang}(t)$, corresponding to linear and angular slip, respectively. A non-slip condition at timestep $t$ is therefore defined as the logical complement of either slip indicator:
    \begin{equation}
        D(t) = \lnot D_{lin}(t) \land \lnot D_{ang}(t)
    \end{equation}
    where $D(t)=1$ denotes that the vehicle is not slipping in either the linear or angular domain. 
    
    The estimated friction coefficient is then defined as the maximum traction coefficient observed over all no-slip timesteps:
    \begin{equation}
        \hat{\mu} = \max_{t \,:\, D(t)=1} \rho(F_w(t))
    \end{equation}
    This ensures that the reported value is consistent with the formal definition of the friction coefficient as the upper bound of traction capability.

    \begin{figure}
        \centering
        \includegraphics[width=0.4\textwidth]{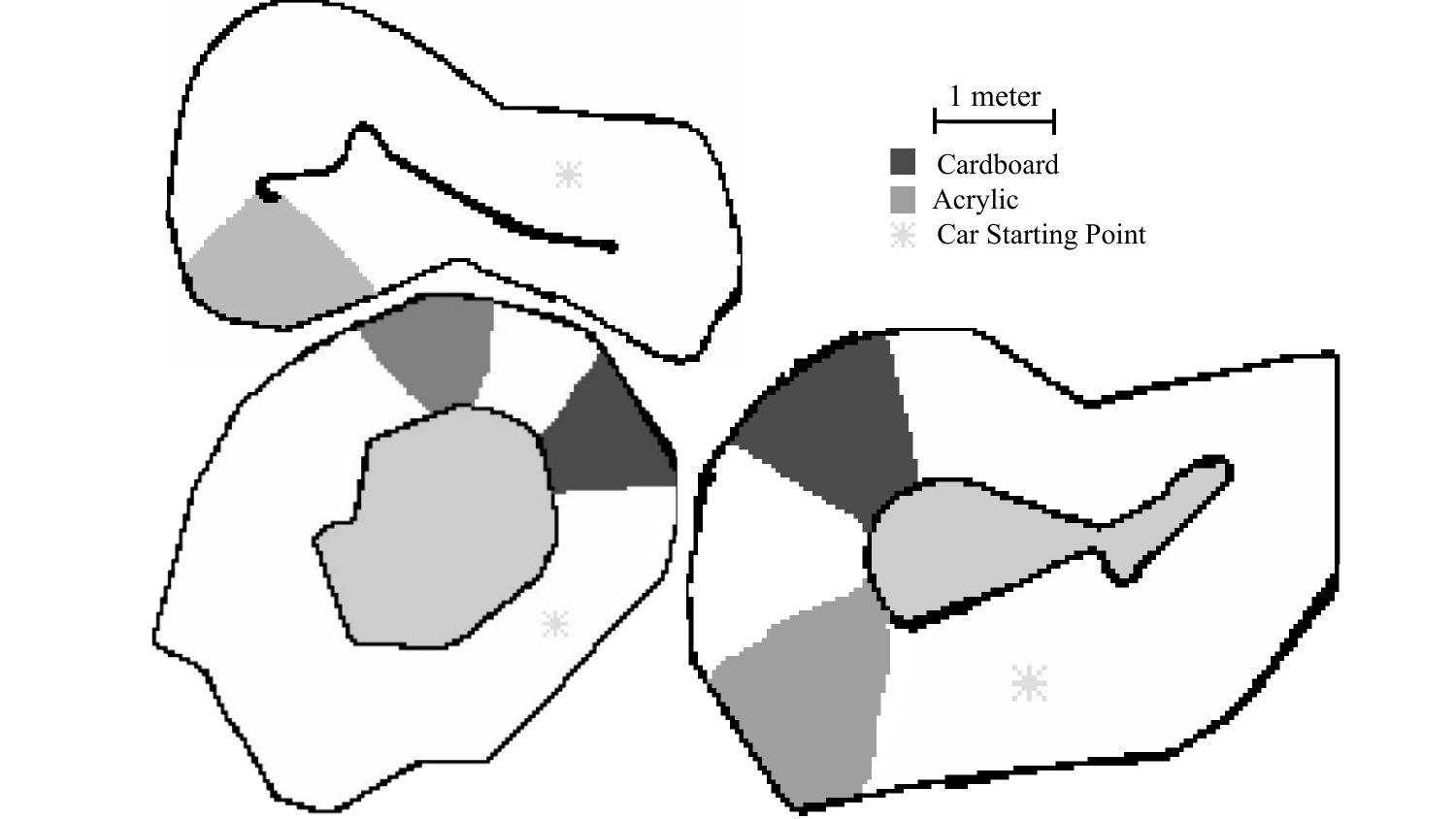}
        \caption{Map of the first, second, and third closed-loop tracks used for testing.}
        \label{fig:myimage3}
        \vspace{-3mm}
    \end{figure}

    \section{Experimental Evaluation}\label{sec:results}
    
    We evaluate our approach along three dimensions:
(1) slip detection accuracy,
(2) slip detection delay, and
(3) tire--road friction coefficient (TRFC) stimation accuracy.

    \begin{figure*}[t]
\centering

\begin{minipage}{0.32\textwidth}
    \centering
    \includegraphics[width=\linewidth]{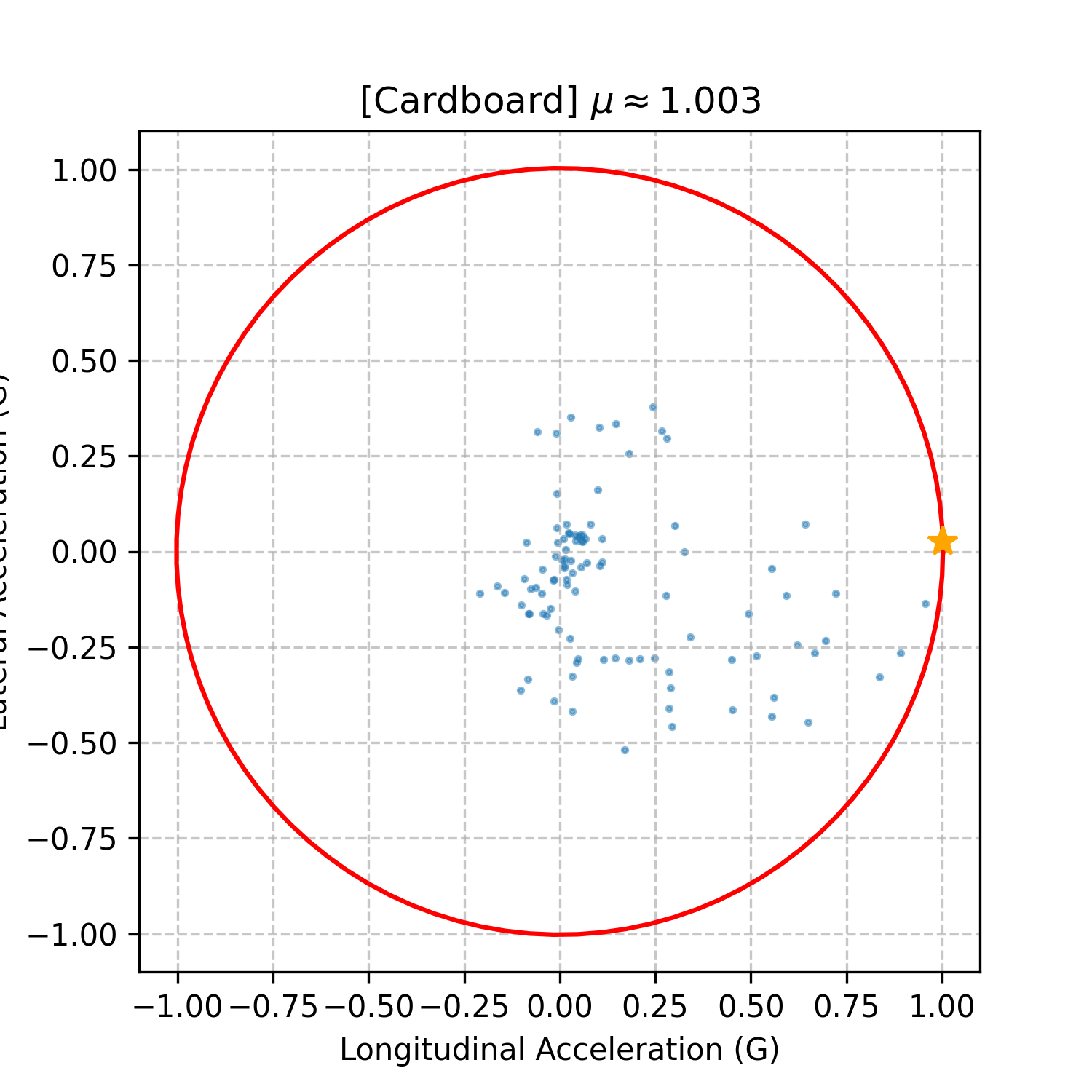}
\end{minipage}
\hfill
\begin{minipage}{0.32\textwidth}
    \centering
    \includegraphics[width=\linewidth]{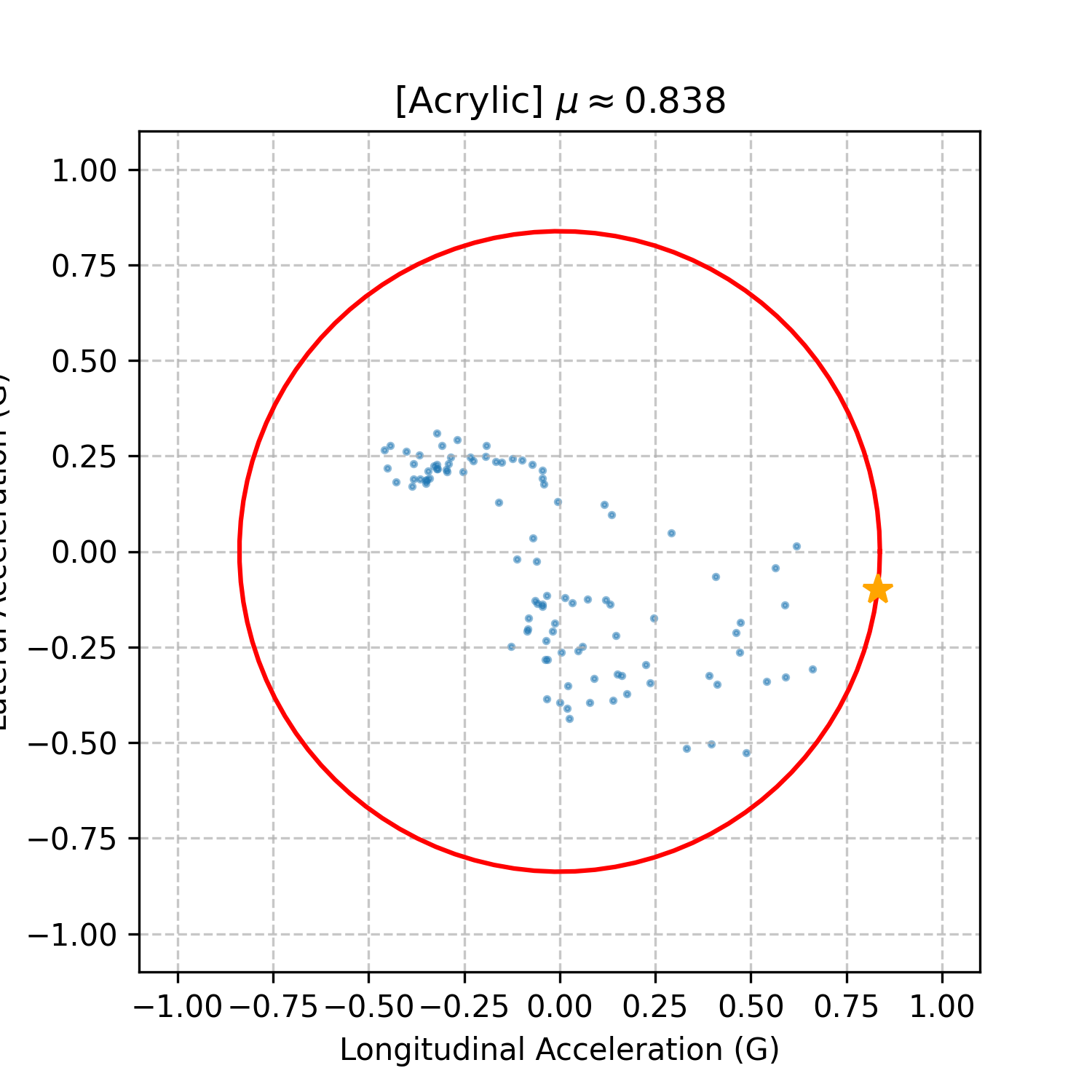}
\end{minipage}
\hfill
\begin{minipage}{0.32\textwidth}
    \centering
    \includegraphics[width=\linewidth]{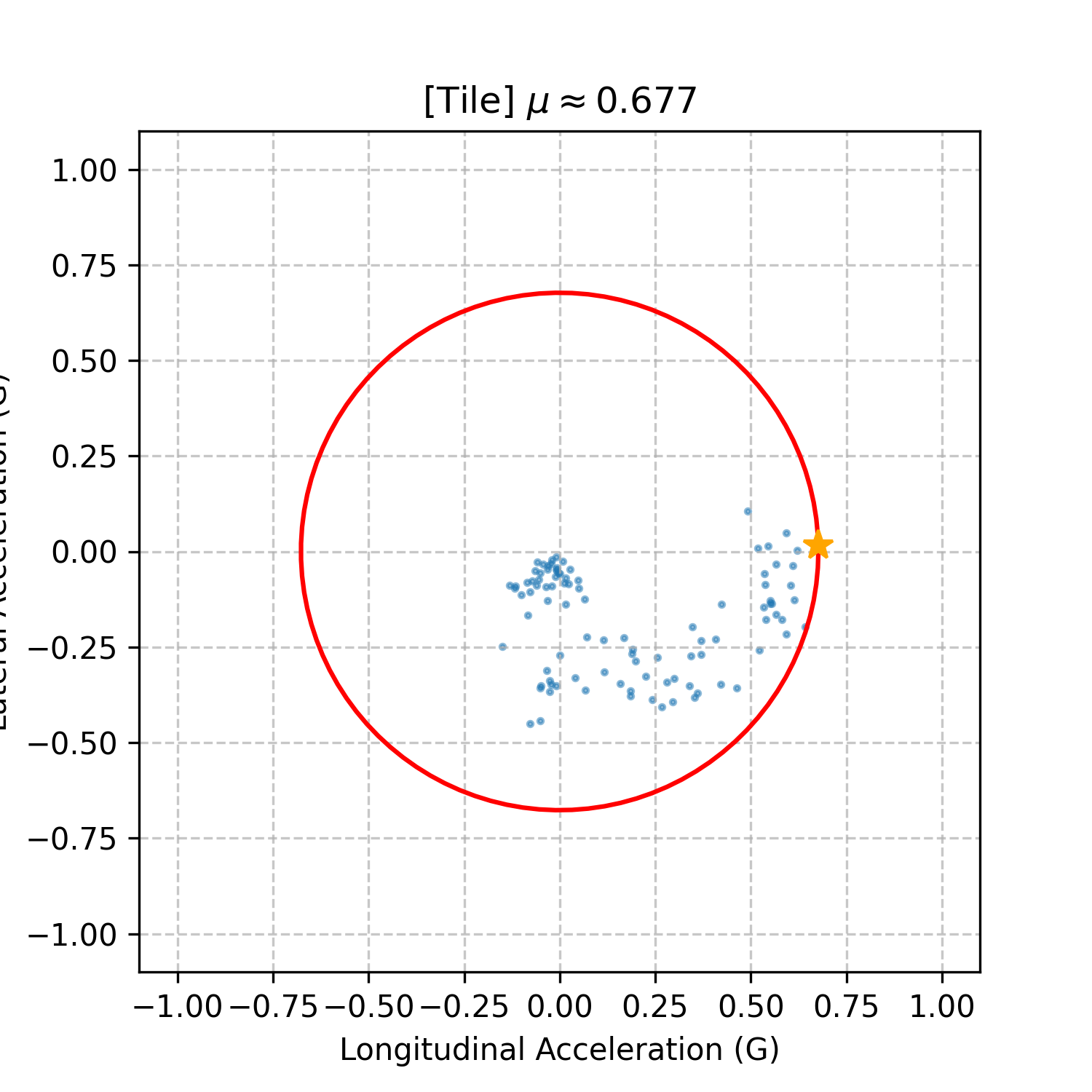}
    
\end{minipage}

\caption{
\looseness=-1
Friction circle visualization for cardboard, acrylic, and tile surfaces. The blue dots denote measured longitudinal and lateral accelerations (in g); the star indicates the maximum traction point corresponding to the estimated friction coefficient $\mu$.}
\label{fig:frictioncircle}
\vspace{-3mm}
\end{figure*}

\subsection{Experimental Setup Overview}

Experiments were conducted using the RoboRacer platform~\cite{okelly2019f1}, a lightweight 1:10-scale autonomous racing vehicle equipped with an IMU and LiDAR and powered by an onboard Jetson Xavier NX. The platform does not include camera-based perception or high-fidelity vehicle instrumentation, and therefore represents a resource-constrained embedded setting. An anonymized implementation of our ROS2 drift detector and the data is available online: \url{https://github.com/Trustworthy-Engineered-Autonomy-Lab/Online-Slip-Detection-Friction-Estimation}

Testing was performed on three closed-loop tracks composed primarily of ceramic tile with embedded sections of cardboard and acrylic to provide three distinct friction regimes. Across all tracks, 56 recorded runs produced 148 labeled slip events spanning all surface types.

\looseness=-1
The three tracks differed in their combinations of tile, cardboard, and acrylic sections, providing two-three distinct friction regimes per track. The specifics are shown in Figure~\ref{fig:myimage3}.

    \subsection{Data Processing}\label{sec:dataproc}

\looseness=-1
    \paragraph*{Data Labeling}
    For each track, ROS2 bags were recorded together with synchronized third-person video. Slip events were manually labeled by aligning the video and bags using the steering-release timestamp and marking the first visible lateral deviation of the vehicle. Although this visual criterion may slightly lag the true physical onset of slip, it provides a consistent and repeatable labeling procedure.

    \paragraph*{Threshold Cross-validation}
    
    To avoid overfitting to specific runs in our evaluation, we perform \textit{K-fold cross-validation} for the detection thresholds $\delta_{lin}$ and $\delta_{ang}$ using the recorded dataset across all tracks, which consists of 56 valid bags. We used $K=5$ folds. Cross-validation was performed independently on each track to ensure that the detection thresholds were not biased toward specific runs or surface configurations.

    In each fold, the training bags were used to set the thresholds; specifically, we used 12 training and 4 testing bags per fold for the 16-bag track, and 16 training and 4 testing bags per fold for the 20-bag tracks. Thresholds were set as two standard deviations above the means of our heuristics $|v_x - \hat{v}_x|$ and $|\omega_\psi - \hat{\omega}_\psi|$. This process ensured that every bag was used exactly once for testing and 4 times for training within its respective track. The testing data from all folds were then used to evaluate slip detection accuracy, detection delay, and friction estimation accuracy across all tracks and surface conditions.

    \subsection{Ground-Truth Friction Coefficient Measurement}

    To determine ground-truth TRFC values on each surface, we conducted a total of 45 independent trials using a force gauge at the vehicle’s center of mass to pull the vehicle until it slipped. For each of the three surfaces, five repeated pulls were applied in each of three directions: lateral, longitudinal, and diagonal ($45^\circ$). For the diagonal direction, pulls were applied from the front of the vehicle while alternating between the two diagonal orientations ($+45^\circ$ and $-45^\circ$) to ensure balanced coverage.
    
    Measured static friction values were approximately
    $\mu \approx 0.69 \pm 0.03$ (tile),
    $\mu \approx 1.02 \pm 0.02$ (cardboard), and
    $\mu \approx 0.84 \pm 0.03$ (acrylic) as shown in gray in Figure~\ref{fig:mugtvsestgraph}. 

    \subsection {Results} 
    \looseness=-1
    \paragraph*{Question 1: Slip Detection Accuracy}
    We evaluated the accuracy of our slip detector by comparing its detected slip events against labeled ground-truth slip times. The slip detector performed \textit{exceptionally well} across our dataset, with a precision of 0.987, a recall of 1.000, and an F1-score of 0.993. That is, our approach detected \textit{every} drift in our dataset (148 events), with no false negatives and only two false positives. Both occurred on track 1 on ceramic tile, near the end of the corresponding bags, where a light contact with the wall contributed to the spurious detections.

    \paragraph*{Question 2: Slip Detection Delay}
    We measured the delay of our slip detection approach by comparing the timestamp of each detection event with the visual onset of slipping recorded in the video of each lap. Our approach demonstrated a \textit{small difference} between the time of drift and the time of detection. The average absolute difference of tile surface detections was $0.407\pm0.534$ s, the average absolute difference of cardboard surface detections was $0.476\pm0.496$ s, and finally the average absolute difference of the acrylic was $0.579\pm0.609$ s.

    \begin{figure}
        \centering
        \includegraphics[width=0.4\textwidth]{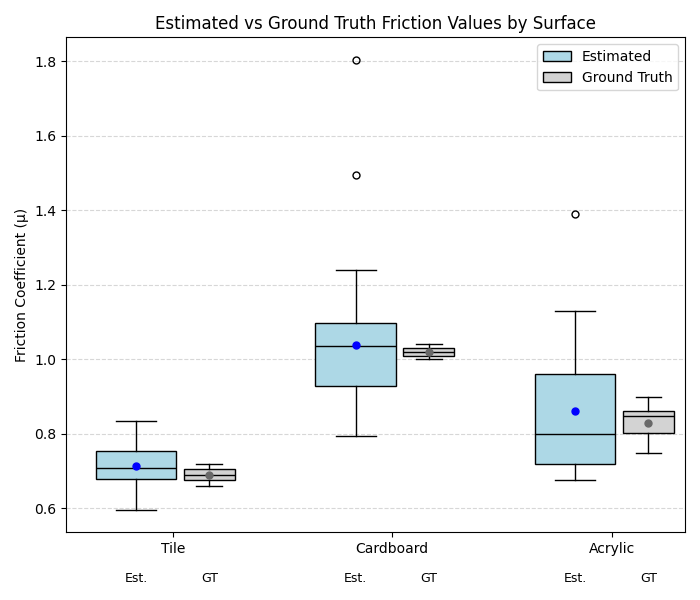}
        \caption{Comparison of ground truth (gray) and estimated (blue) friction coefficients (µ) on tile, cardboard, and acrylic.}
        \label{fig:mugtvsestgraph}
        \vspace{-4mm}
    \end{figure}

    \looseness=-1
    \paragraph*{Question 3: Friction Estimation Accuracy}
    To determine the accuracy of our friction coefficient estimation, we compared the measured ground-truth TRFC values with those produced by our approach. Our estimated values were $\mu \approx 0.714$ on tile, $\mu \approx 1.038$ on cardboard, and $\mu \approx 0.860$ on acrylic. These values were, on average, \textit{accurate} compared to the ground truth values, with a mean absolute error (MAE) of $0.042\pm 0.032$ for the tile surface, $0.116\pm 0.120$ for the cardboard surface, and $0.139\pm 0.097$ for the acrylic surface (visualized in blue in Figure~\ref{fig:mugtvsestgraph}). As expected, our approach showed some variance due to sensor noise. Figure~\ref{fig:frictioncircle} shows an example of our TRFC estimation results on cardboard, acrylic, and tile surfaces. The red circle represents the friction circle, indicating the limit of possible tire force combinations where maximum friction is estimated. The variability of acceleration demonstrates that our detection and estimation approach was tested comprehensively.

\looseness=-1
    Table~\ref{tab:error_compact} contextualizes our experimental results in terms of existing TRFC estimation approaches. The metrics include MAE and the mean relative error (MRE, i.e., MAE normalized by the true TRFC). Note that for the adaptive cubature KF~\cite{wang2020ackf_trfc}, MAE values were not explicitly reported; therefore, we approximated them by digitizing the published plots and extracting the corresponding error values.  Recall that, unlike most model-based approaches, which rely on explicit vehicle or tire models and parameter identification, and unlike learning-based methods that require extensive training data, our approach operates without parametric modeling or prior training, using \textit{only} IMU and LiDAR measurements together with control inputs. Despite this lightweight formulation, the achieved estimation errors remain within a comparable range to filtering-based techniques. In particular, across multiple surface conditions, our method maintains consistent performance without surface-specific tuning or calibration. These results support the central claim of this work: accurate friction estimation can be achieved through a simple, model-free, and computationally efficient framework suitable for embedded onboard deployment.

\begin{table}[t]
    \centering
    \caption{Error comparison for tire--road friction coefficient estimation methods. \textit{MRE} denotes \textit{MAE} normalized by the ground-truth.}
    \label{tab:error_compact}
    \footnotesize
    \setlength{\tabcolsep}{3pt}
    \renewcommand{\arraystretch}{1.05}
    \begin{tabular}{lccc}
    \hline
    Method (paper) & MAE & MRE (\%)\\
    \hline
    Deep-learning based \cite{song2018estimating} & 0.019 & \text{2.77} \\
    \hline
    Refined adaptive KF \cite{li2024adaptiveSTKF} & 0.142 & \text{17.75} \\
    \hline
    
    & 0.02 & 3.6 \\
    & 0.01 & 2.3 \\
    Predictive ELSD Control (multiple surfaces) \cite{woo2024elsd} & 0.02 & 8.3 \\
    & 0.01 & 5.5 \\
    & 0.02 & \text{2.2} \\
    \hline
    Adaptive cubature KF (approximation) \cite{wang2020ackf_trfc} & 0.127 & \text{15.9} \\
    \hline
    Wavelet transform and neural network \cite{sadeghi2022maximum} & 0.043 & 5.30 \\ \hline
    
    & 0.042 & 6.09 \\ 
    Ours (multiple surfaces) & 0.116 & 11.37 \\
    & 0.139 & 16.45 \\
    \hline
    \end{tabular}
\end{table}

    \section{DISCUSSION AND CONCLUSION}\label{sec:conclusion}

    \looseness=-1
    This paper presented a method for onboard slip detection and friction coefficient estimation in real-time. The approach identifies slip by comparing commanded velocities from control actions with measured velocities estimated via LiDAR-based EKF, declaring slip when discrepancies exceed tuned thresholds. The tire-road friction coefficient is then estimated directly from IMU-measured accelerations with the maximum observed value during non-slip intervals reported as the effective coefficient. Experiments on a 1:10-scale autonomous racing car over tile and cardboard surfaces demonstrated that onboard estimates closely matched ground-truth measurements obtained with a force gauge, with differences consistently within 5-10\%. These results highlight the potential of lightweight, low-cost methods to capture frictional properties critical to reliable driving, without reliance on explicit tire models or large training datasets.
    
    Nevertheless, several limitations must be acknowledged. The experiments were restricted to three surfaces and a limited number of trials, which constrains the generalizability of the results. Environmental factors such as humidity, wear of the surfaces, and variation in the manual pulling force applied to the vehicle’s center of mass during ground-truth measurements were not fully controlled, potentially influencing repeatability. Additionally, the current method assumes largely isotropic surfaces and enforces simplifying assumptions introduced in Section~\ref{sec:problem} --- including planar dynamics, negligible aerodynamic drag and rolling resistance, and single-track approximation. These assumptions may not hold for significantly textured or anisotropic terrains encountered in real-world deployments.
    
    Future work will focus on extending validation beyond controlled indoor surfaces to more diverse and dynamically changing conditions, such as outdoor asphalt and wet or low-adhesion environments. Incorporating additional vehicle dynamics effects, including aerodynamic drag and load transfer, would further improve estimation accuracy during higher-speed or aggressive maneuvers. Another direction is improving slip onset localization using sensor-driven criteria (encoder-based) rather than visual labeling, enabling more precise delay analysis. Finally, hybrid extensions that combine the present physics-based framework with lightweight learning components may enhance generalization across unseen surfaces while preserving the interpretability and computational efficiency of the current approach.
    
    This study provided an initial validation of onboard friction estimation techniques, showing strong alignment with external ground-truth methods. While the scope was intentionally narrow, the findings demonstrate that even lightweight onboard approaches can yield reliable surface characterizations, laying the groundwork for more comprehensive future investigations.

    \section*{ACKNOWLEDGMENT}

    We gratefully acknowledge the support of the University Scholars and AI Scholars programs at the University of Florida. We would also like to thank Anthony Joannides, Gage Pollard, and Zhongzheng Ren Zhang for their help in setting up and operating racing cars.

    This research was supported in part by the NSF award CNS 2513076.
    Any opinions, findings, conclusions, or recommendations expressed in this material are those of the authors and do not necessarily reflect the views of the National Science Foundation (NSF) or the US Government.

    

	\bibliographystyle{IEEEtran}
	\bibliography{root} 
    
\end{document}